\pdfoutput=1

\documentclass[11pt]{article}

\usepackage[]{acl}

\usepackage{times}
\usepackage{latexsym}

\usepackage[T1]{fontenc}

\usepackage[utf8]{inputenc}
\usepackage{microtype}
\usepackage{multirow}
\usepackage{setspace}
\usepackage{inconsolata}
\usepackage{booktabs,multirow}
\usepackage{adjustbox}
\usepackage{subfig}
\usepackage{hyperref}
\usepackage{tcolorbox}
\usepackage{xcolor}
\usepackage{amsmath}
\usepackage{mdframed}
\usepackage{tabularx}
\usepackage{longtable}
\usepackage[ruled,linesnumbered]{algorithm2e}
\usepackage{algpseudocode}
\usepackage{amssymb}
\usepackage{multirow}
\usepackage{hhline}
\usepackage{tabularx}
\usepackage{pdfpages}
\usepackage{authblk}

\title{EIPE-text: Evaluation-Guided Iterative Plan Extraction for Long-Form Narrative Text Generation}

\author[]{\bf Wang You\thanks{\ \ Equal contribution} \hspace{0.1em}\thanks{\ \ Work done during internship at Microsoft Research Asia.} \quad Wenshan Wu$^*$\thanks{\ \ Corresponding author: wenswu@microsoft.com}   \quad Yaobo Liang$^*$ \quad Shaoguang Mao\quad \\ \bf Chenfei Wu \quad  Maosong Cao$^\dag$\quad Yuzhe Cai$^\dag$\quad Yiduo Guo$^\dag$\quad Yan Xia\quad Furu Wei \quad Nan Duan}

\affil[]{Microsoft Research Asia}

\begin{document}
\maketitle

\begin{abstract}

Plan-and-Write is a common hierarchical approach in long-form narrative text generation, which first creates a plan to guide the narrative writing. Following this approach, several studies rely on simply prompting large language models for planning, which often yields suboptimal results. In this paper, we propose a new framework called Evaluation-guided Iterative Plan Extraction for long-form narrative text generation (EIPE-text), which extracts plans from the corpus of narratives and utilizes the extracted plans to construct a better planner. EIPE-text has three stages: plan extraction, learning, and inference. In the plan extraction stage, it iteratively extracts and improves plans from the narrative corpus and constructs a plan corpus. We propose a question answer (QA) based evaluation mechanism to automatically evaluate the plans and generate detailed plan refinement instructions to guide the iterative improvement. In the learning stage, we build a better planner by fine-tuning with the plan corpus or in-context learning with examples in the plan corpus. Finally, we leverage a hierarchical approach to generate long-form narratives. We evaluate the effectiveness of EIPE-text in the domains of novels and storytelling. Both GPT-4-based evaluations and human evaluations demonstrate that our method can generate more coherent and relevant long-form narratives. Our code will be released in the future.
\end{abstract}

\begin{figure*}[!t]
    \centering
    \includegraphics[width=\linewidth]{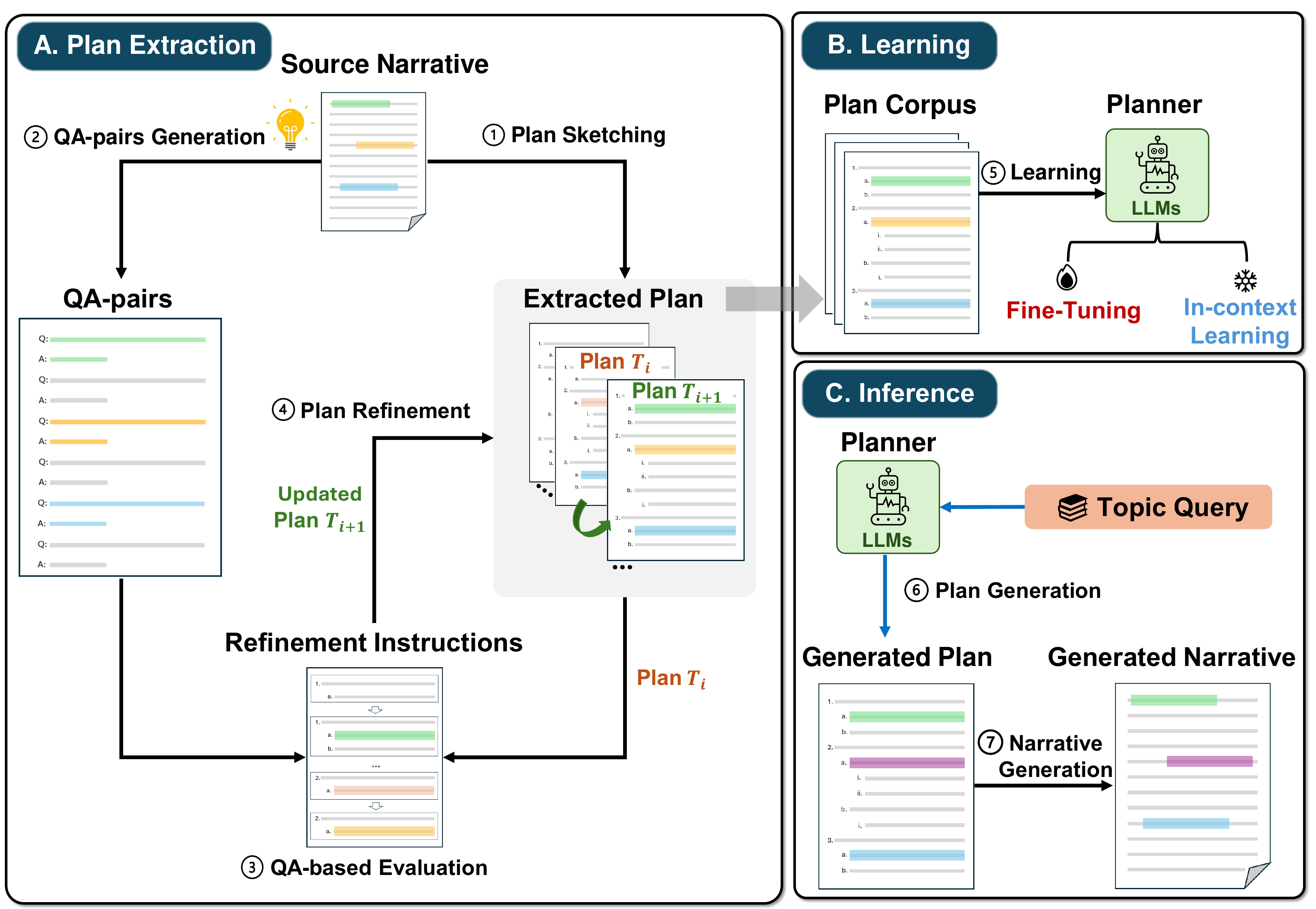}
    \caption{\textbf{A Comprehensive Visual Overview of the EIPE-text Framework.} The \textbf{Plan Extraction} stage starts with \textit{Plan Sketching}, where an initial plan is generated using an LLM. Then, in the \textit{QA-pairs Generation} step, a set of QA-pairs is created to evaluate the plan. \textit{QA-based Evaluation} step evaluates the plan through question answering and generates refinement instructions. In the \textit{Plan Refinement} step, it iteratively improves the plan based on the instructions until it passes the evaluation. Plans are then used to construct a plan corpus for the planner in the \textbf{Learning} stage. Finally, in the \textbf{Inference} stage, the planner generates a plan, and the narrative is generated from the plan.}
    \label{fig:illustration}
\end{figure*}

\section{Introduction}
Large language models have made impressive strides in text generation, performing well in tasks such as machine translation, summarization, and chat \citep{chang2023survey}\citep{bubeck2023sparks}. However, generating long-form narrative remains a challenging task, especially when it comes to maintaining coherence over long ranges and ensuring relevance to an initial premise. This is particularly crucial for applications such as scriptwriting, novels, business reports, journalism, among others. 

Human writers often create a plan or outline before beginning to write a narrative, which helps maintain a coherent and logical progression throughout the narrative. Inspired by this, a hierarchical generation approach has been used in many works, such as Re3\citep{yang-etal-2022-re3}, DOC\citep{yang-etal-2023-doc}, and recurrentGPT\citep{zhou2023recurrentGPT}. 
These works mainly focus on how to generate the full narrative based on a plan and only generate the plan by simply prompting a large language model. However, the planning ability of LLMs is not good enough and requires significant prompting engineering work. Additionally, it is challenging to adapt these models to a specific domain or style of long-form narrative.

To address these limitations, we propose the Evaluation-Guided Iterative Plan Extraction for Long-Form Narrative Text Generation (EIPE-text) framework. EIPE-text leverages a learned planner with enhanced domain expertise to generate a high-quality plan, as illustrated in figure \ref{fig:illustration}. Specifically, EIPE-text consists of three stages: plan extraction, learning, and inference. In the plan extraction stage, we iteratively extract and improve plans from collected narrative corpus to construct a plan corpus for planner learning. To evaluate the quality of extracted plans and the alignment between plans and source narratives, we adopt a QA-based self-evaluation mechanism, leveraging the reading comprehension capabilities of LLMs. Based on evaluation results, we generate detailed refinement instructions to iteratively improve the plan. In the learning stage, we build a better planner by fine-tuning with the plan corpus or in-context learning with examples in the plan corpus, which enhances the ability to generate high-quality plans. During the inference stage, we first generate the plan and then further generate narratives based on the plan.

We evaluated the effectiveness of EIPE-text in the domain of novels and storytelling and found that both the fine-tuning based and in-context learning based planners outperform the baselines. Human evaluation also shows that the results of EIPE-text were more coherent and relevant than those of current state-of-the-art models.

Our contributions can be summarized as follows:
\begin{itemize}
    \item We propose a new framework, EIPE-text, which automatically extracts high-quality plans from narrative corpus and learns better planners for long-form narrative text generation. This framework can be generalized to all domains.
    \item We propose a QA-based evaluation method to automatically evaluate plans and generate detailed instructions to improve the plan based on evaluation results. This QA-based evaluation provides more specific and actionable results than simply leveraging GPT to compare two outputs or provide a score ~\cite{liu2023gpteval}. 
    \item We demonstrate the effectiveness of our model in the novel and storytelling domains, and we will release the code for future research.

\end{itemize}

\section{Method}
\label{sec:2}

Our methodology contains three stages: plan extraction, learning, and inference. The entire process is shown in figure \ref{fig:illustration}. During the plan extraction phase, plans are extracted from each narrative within the corpus. These extracted plans are then compiled to construct the plan corpus. By relying on the constructed planning corpus, planner can learn to generate high-quality plans. In the inference stage planner generates a better plan and a narrative will be generated from the plan.

The plan extraction stage contains plan sketching, QA-pairs generation, QA-based evaluation, and plan refinement. Initially, we create a tree-structured plan using the LLM in the plan sketching step. Next, during the QA-pairs generation phase, we generate a set of QA-pairs, with each pair corresponding to a distinct part within the source narrative. These QA-pairs serve as an evaluation metric for the plan. The QA-based evaluation step evaluates the plan by question answering. For each incorrect QA-pair, we generate corresponding instructions to modify the relevant part of the plan. In the plan refinement step, we integrate these instructions received in previous steps to update the plan. We repeat steps 3 and 4 until the extracted plan passes the evaluation.


In the learning stage, We leverage the plan extracted in the first stage to train an LLM planner. To achieve this, we utilize two strategies: finetuning, as well as in-context learning. These strategies contribute to generating high-quality plans for the given topic.

The inference stage contains two steps: plan generation and narrative generation. Firstly, the planner takes the topic as input and generates a corresponding plan. Secondly, the narrative will be generated in the narrative generation step.

\begin{algorithm*}
\SetKwInOut{Input}{Input}\SetKwInOut{Output}{Output}
\caption{Plan Extraction Algorithm}
\label{alg:plan_extraction}

\Input {$\mathcal{C}_n=\{n_1, n_2, ..., n_m\}$} 
\Output {$\mathcal{C}_p=\{p_1, p_2, ..., p_m\}$}
$\mathcal{C}_p \leftarrow \varnothing$ 

\For{$i\leftarrow 1$ \KwTo $m$}{

    $p^0_i \leftarrow \text{plan\_sketching}(n_i)$ 
    
    $\mathcal{C}_q \leftarrow \text{qa\_pairs\_generation}(n_i)$ \Comment{$\mathcal{C}_q=\{q_1, q_2, ..., q_k\}$ \textbf{questions set}}   

    $t\leftarrow 0$ \Comment{$t$ \textbf{refinement time step}}

    \While{\textbf{not pass\_evaluation}$(p^t_i,  \mathcal{C}_q)$} {
        $\mathcal{C}_i \leftarrow \text{qa\_based\_evaluation}(p^t_i,  n_i,  \mathcal{C}_q)$ \Comment{$\mathcal{C}_i=\{i_1, i_2, ..., i_l\}$ \textbf{refinement instructions set}}
        
        $p^{t+1}_i \leftarrow \text{plan\_refinement}(p^t_i, \mathcal{C}_i)$ 

        $t\leftarrow t + 1$
    }
    $\mathcal{C}_p \leftarrow \mathcal{C}_p \cup p^t_i$ 
}
\end{algorithm*}

\subsection{Plan Extraction}
Formally, we have a corpus of narrative $\mathcal{C}_n=\{n_1, n_2, ..., n_m\}$. The plan extraction stage extracts a plan $p_i$ for each narrative $n_i$. The extraction results are compiled to a plan corpus $\mathcal{C}_p=\{p_1, p_2, ..., p_m\}$. We illustrated the process of plan extraction in algorithm \ref{alg:plan_extraction}.

\paragraph{Plan Sketching.}
For each narrative, we use LLM to extract a tree-structured plan, which serves as the plan sketch. The detailed LLM prompt can be found in appendix \ref{sec:plan_sketching_prompt}. The plan is in a tree structure and the content of each \textbf{node} is the summarization of the corresponding section, subsection, and so forth. We show an example of a plan sketch in figure \ref{fig:plan_comparison}.

\paragraph{QA-pairs Generation.}
For each narrative, we generate a set of QA-pairs, with each pair corresponding to a different segment of the narrative. These QA-pairs can be utilized to evaluate whether the plan includes all aspects of the narrative. Each QA-pair is formulated as a multiple-choice problem, comprising one question, multiple candidate answers, and multiple correct answer indices. The number of QA-pairs is proportional to the length of the narrative. To ensure the quality of the generated QA-pairs, we employ another LLM to answer these questions based on the original text, filtering out any incorrectly answered pairs. The guidelines for this process can be found in appendix \ref{sec:qa_generation_guideline}.

\paragraph{QA-base Evaluation.}
We evaluate a plan using QA-pairs and provide detailed refinement instructions for refining the plan further. Specifically, we utilize LLM to answer questions based on the plan. For each incorrect question, we generate an instruction to modify the plan so that the question can be correctly answered. The modification instruction can be one of the following: (1) \textbf{add}, which inserts a missing node into the plan; (2) \textbf{modify}, which alters the content of a node; (3) \textbf{adjust}, which relocates a node to another level of the tree, thereby altering the tree's structure. Detailed refinement instructions enable LLM to make precise improvements to specific parts of the plan.
\paragraph{Plan Refinement.}
In this step, we incorporate the instructions generated in the previous step to improve the plan. Ideally, we should apply the changes one by one. In order to improve efficiency, we instruct the LLM to apply all instructions simultaneously. However, the refinement instructions generated by LLM may not always address the incorrect questions. Therefore, we iteratively perform the refinement instructions generation and plan refinement steps until the new plan can pass the QA-based evaluation. This process ensures that the final plan has addressed all the identified errors and meets the desired quality standards.

While LLM possesses a self-improving ability and can refine the plan through simple prompting, the quality of the improvement results may still not be good enough or even worse. Our QA-based evaluation, on the other hand, can identify specific errors in the plan and provide refinement instructions in the form of instructions to enhance the plan. This approach can achieve better refinement performance.

\subsection{Learning}

During the learning phase, we implemented two methods to enhance the performance of the planner: the in-context learning method and the fine-tuning method. 

The in-context learning method improves the planner by selecting representative demonstration examples from the plan corpus. By selecting different demonstration examples, the fixed LLM can quickly adapt to specific domains or styles.

On the other hand, the fine-tuning method can further improve the planner's ability by training it on all plan corpus. This method leverages all the data in the plan corpus and enables the planner to adapt to multiple domains simultaneously. 

\subsection{Inference} 
The inference stage comprises two steps: plan generation and narrative generation.
\paragraph{Plan Generation.}
In this step, the planner takes the chosen topic as input and produces a corresponding plan. The planner constructs a well-structured plan that outlines the key elements and sections to be covered in the ensuing narrative.
\paragraph{Narrative Generation.}
The narrative is generated from the generated plan in this step. This narrative seamlessly integrates the content outlined in the plan, ensuring that the resulting narrative is not only logically organized but also rich in detail and context. The final narrative is a well-rounded piece of long-form narrative that effectively conveys the information related to the chosen topic.

\subsection{Discussion}
In this section, we will discuss how EIPE-text works. Here is our analysis:

Let $q$ be the premise query. The probability of desired output based on premise query $p(n|q)$ could be rewritten as 
\begin{equation}
P(n|q)=P(p|q)P(n|p) \label{eq1}
\end{equation}
When plan $p$ is of high quality, $P(n|p)$ will be high. So as $P(p|q)$ increases, $P(n|q)$ increases too. Our framework EIPE-text actually increases $P(p|q)$.

Besides, the process of plan refinement in figure \ref{fig:illustration} could be understood as Reinforcement Learning(RL), LLM gets observation from answering the question, and then obtains refinement instructions according to the true or false case. After obtaining refinement instructions, LLM changes the original state to the new state i.e. revise plan. After many interactions with the "environment", the "state" will be iterated to a suitable "state" that can be used to improve $P(p|q)$.

To practically exemplify the effectiveness of EIPE-text, we conducted a case study of plan generation through in-context learning with one demonstration. A detailed exploration of this case is provided in the Appendix \ref{appendix:plan_generate} for interested readers.

\begin{table*}[ht]
\centering

\begin{tabular}{lcccc} 
\toprule
\textbf{Dataset} & \textbf{Train Size} & \textbf{Test Size} & \textbf{Avg Length} & \textbf{Max Length} \\
\midrule
TED Talk & 2468 & 130 & 2078 & 9044 \\
Novel & 1292 & 120 & 3741 & 14493 \\
\bottomrule
\end{tabular}
\caption{Comprehensive Dataset Information for TED Talk and Novel.}
\label{tab:1}
\end{table*}

\begin{table}[]
\small
\centering
\begin{tabular}{@{}lccc@{}}
\toprule
Novel genres    & \multicolumn{3}{c}{Overall(human)}           \\ \midrule
$\sim$4500words & Interesting      & Coherent   & Relevant      \\ \midrule
EIPE-text (in-context)         & 56.7 & \textbf{64.2} & \textbf{75.8} \\
recurrentGPT    & \textbf{60.0}          & 59.2          & 62.5          \\ \bottomrule
\end{tabular}
\caption{Novel Human Evaluation Results. Pair-wise comparison using human evaluation of EIPE-text with recurrentGPT for 120 novels of different genres. Results never mix numbers from different comparisons}
\label{tab:2}
\end{table}

\begin{table}[]
\small
\centering
\begin{tabular}{@{}lccc@{}}
\toprule
Novel genres    & \multicolumn{3}{c}{Overall(automatic)}           \\ \midrule
$\sim$4500words & Interesting      & Coherent   & Relevant      \\ \midrule
EIPE-text (in-context)         & 55.0 & \textbf{84.2} & \textbf{92.5} \\
recurrentGPT    & \textbf{58.3}          & 65.8          & 84.2          \\ \bottomrule
\end{tabular}
\caption{Novel GPT4 Evaluation Results. Pair-wise comparison using GPT-4 evaluation of EIPE-text with recurrentGPT for 120 novels of different genres. Results never mix numbers from different comparisons}
\label{tab:3}
\end{table}

\begin{table*}[]
\centering
\begin{tabular}{@{}llcc@{}}
\toprule
setting A                 & setting B            & A Win Ratio & B Win Ratio   \\ \midrule
LLaMA raw planner & EIPE-text (finetune)   & 6.2         & \textbf{93.8} \\
GPT4 raw planner & EIPE-text (in-context) & 22.5        & \textbf{75.2} \\ \bottomrule
\end{tabular}
\caption{TED Talk Automatic Evaluation Results. Pair-wise comparison using GPT-4 evaluation of EIPE-text with baselines for 130 TED talk transcripts. Results in different comparisons are not comparable with each other.}
\label{tab:4}
\end{table*}


\section{Experiments}
In this section, we compare EIPE-text in novels and storytelling generation with the baselines. All experiments show that EIPE-text is better than the baselines, verifying the effectiveness of our framework.

\subsection{Setup}
\label{sec:setting}
For plan extraction stage, we use Azure Openai GPT-4 as our experimental LLM. And for inference stage, we use the planner to generate a plan to further generate the narrative. It should be emphasized that we did not intentionally implement the narrative generation, but modified it based on recurrentGPT, as described in the appendix \ref{sec:appendixb}. \textbf{For all the settings mentioned in the following section, unless special emphasis, they adhere to the description provided above.}

\subsection{Novel}
\subsubsection{Dataset} Novels are long-form narratives that include intricate plots, and rich character development. The model needs to maintain consistency in plots and character development and generate interesting stories. We use the data collected from Project American Literature\footnote{\url{https://americanliterature.com/short-stories}}, Writing Prompts\footnote{\url{https://blog.reedsy.com/creative-writing-prompts/}} and etc. Then we aggregate a training dataset containing total 1292 stories. Besides, we collected 120 prompts as a test set from Writing Prompts, which cover six genres. The more information about this dataset is shown in table \ref{tab:1}. 
\subsubsection{Setting} 
\paragraph{EIPE-text (in-context)} For learning stage, we use the \textit{text-embedding-ada-002}, to obtain text embeddings of plan corpus. These embeddings will then be utilized in conjunction with the \textit{k-means} algorithm for cluster purposes. We use \textit{k-means} getting 20 clustering centroids as demonstrations to learn a planner and use the planner during comparing with baselines. 
\subsubsection{Baselines}
\paragraph{recurrentGPT} A language-based simulacra of the recurrence mechanism in RNNs that uses language-based components and defines a recurrent computation graph via prompt engineering.

It is worth mentioning that we are not directly comparing with Re3 and DOC, because recurrentGPT is already way ahead of these methods.
\subsubsection{Metric}
Our evaluation employs a pairwise comparison metric. We report results individually for each pairwise comparison between EIPE-text and each baseline, never mixing numbers from different comparisons following Re3 \citep{yang-etal-2022-re3}. We show the criteria as outlined in \citep{yang-etal-2023-doc} for novel as following:
\begin{itemize}
    \item \textbf{Interesting}: An interesting novel captivates the reader's attention, engages them emotionally, and holds their interest throughout.
    \item \textbf{Coherent}: A coherent novel follows a logical and consistent plot-line without significant gaps or inconsistencies.
    \item \textbf{Relevant}: Faithful to the initial premise.
\end{itemize}

\paragraph{Automatic Evaluation}
For automatic evaluation, we employed GPT-4 to assess various aspects of the generated narrative. GPT-4 automatic evaluation is highly affected by the order and unstable, so all metrics are judged by GPT4 with a premise, aforementioned criteria and two corresponding stories in random order. We also use majority voting system to evaluate each criterion of each pair. The evaluation prompt for novel can be found in appendix \ref{sec:appendixc1}.
\paragraph{Human Evaluation}
In order to ensure impartial and high-quality evaluations, we collaborated with third-party data annotators. Each generated data pair, comprising novels A and B presented in random order, underwent meticulous evaluation by three distinct annotators. These annotators possess proficient English language skills and were provided with explicit instructions to evaluate and deliver judgments on the superiority between novel A and novel B, or if they are indistinguishable, specifically in relation to the aforementioned criteria.
\subsubsection{Result} 
We show the experiment results of novels in table \ref{tab:2} and table \ref{tab:3}. As we can see from the table, EIPE-text shows an advantage in coherence and relevance in both human and automatic evaluation. Although the human evaluation is less interesting (3.3\%), the improvement of coherence (5.0\%) and relevance (13.3\%) are significant. The same trend can be seen in automatic evaluation, it is less interesting than recurrentGPT(3.3\%), but coherent (18.4\%) and relevant (8.3\%) are significantly higher. These results indicate that EIPE-text improves the overall quality of generated narrative, and also indicate that automatic evaluation and human evaluation have certain relevance.

\subsection{Strorytelling}
\subsubsection{Dataset} TED Talks \footnote{\url{https://www.ted.com/talks}} are influential presentations that cover a wide range of topics. They are known for their engaging narratives, concise structure, and powerful messages, which can be challenging to generate for both models and humans. We use the data collected by Kaggle \footnote{\url{https://www.kaggle.com/datasets/rounakbanik/ted-talks}}. The training dataset aggregates 2,468 TED Talks spanning the years 1984 to 2016. In addition, we have curated 130 TED Talk transcripts post-2021 as our testing datasets as shown in table \ref{tab:1}. 
\subsubsection{Setting} 
\paragraph{EIPE-text (in-context)} For learning stage, text embeddings obtained using \textit{text-embeddings-ada-002} are used for clustering together with the \textit{k-means} algorithm. Then we use 20 clustering centroids as demonstrations to learn a planner. 
\paragraph{EIPE-text (finetune)} We finetune the open source LLM, LLaMA \citep{touvron2023llama}, using the plan corpus and use it as planner during learning stage. Specially, we finetune LLaMA-7B using lora\citep{hu2022lora}.
\subsubsection{Baselines}
\paragraph{GPT4 raw planner} In this setup, planner is GPT4 zero-shot whose ability to plan depends entirely on its native capabilities. After the planner generates the plan, narrative generation follows the same way as the inference stage in \ref{sec:setting}
\paragraph{LLaMA raw planner} similar to GPT4 raw planner, but the planner is untrained LLaMA. 

\subsubsection{Metric}
We only adopt automatic evaluation in storytelling generation. The evaluation criteria were tailored to specific domain to ensure relevant and accurate assessments, so we use other criteria for storytelling:
\begin{itemize}
\item \textbf{Coherent}: The talk should have a clear structure, smooth transitions, and a strong conclusion for easy comprehension and a consistent theme.
  
\item \textbf{Interesting}: It should use storytelling and examples to engage the audience, maintaining their curiosity throughout.
  
\item \textbf{Relevant}: The topic should be timely, address current issues, and offer fresh insights, not just repeat existing information.
  
\item \textbf{Inspiring}: The talk should convey passion, present innovative ideas, and encourage the audience to think differently or take action.
\end{itemize}

It should be emphasized that we only use majority voting system to evaluate each pair for all criteria, instead of evaluating each criterion of each pair. The evaluation prompt for storytelling can be found in appendix \ref{sec:appendixc2}

\subsubsection{Results}

We show the experiment result of storytelling domain on TED Talk in table \ref{tab:4}. Under the finetune setting, EIPE-text far outperforms LLaMA raw planner (87.6\%). Also under setting B, EIPE-text is significantly outperform the GPT4 raw planner (52.7\%). EIPE-text either using a finetune base planner or using in-context learning based planners is well ahead of the LLM itself.

\begin{figure*}[!t]
    \centering
    \includegraphics[width=\linewidth]{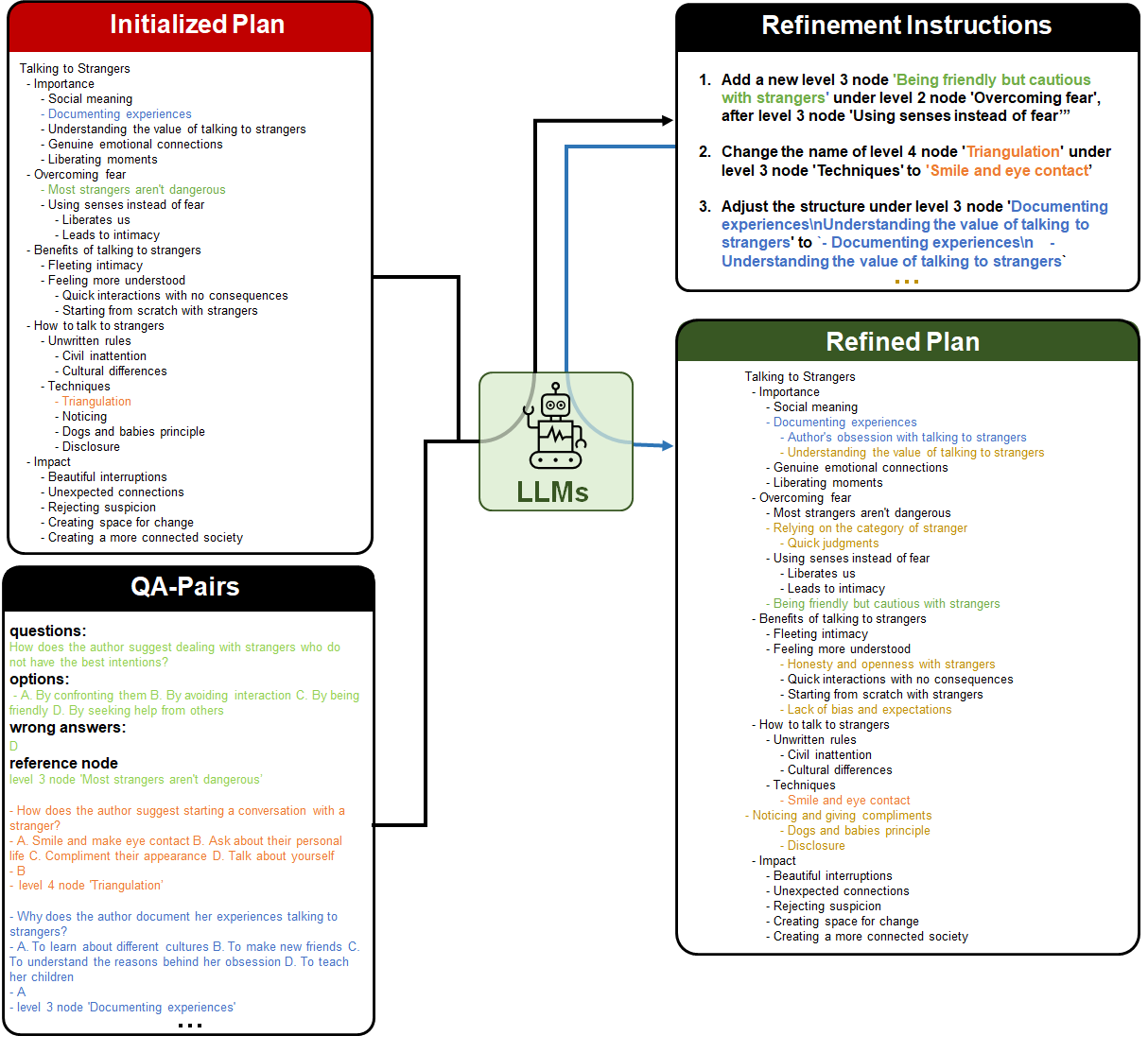}
    \caption{An Example of the Plan Refinement Process.}
    \label{fig:plan_comparison}
\end{figure*}

\begin{table*}[]
\centering
\small
\renewcommand{\arraystretch}{1.3}
\begin{tabular}{llcc}
\hline
\textbf{A}                 & \textbf{B}                    & \textbf{A Win Ratio} & \textbf{B Win Ratio} \\ \hline
\multicolumn{4}{l}{\textbf{Different Demonstration Number}}                                              \\ \hline
20-shot cluster-based planner & 5-shot cluster-based planner    & \textbf{70.9}        & 26.8                 \\ \hline
\multicolumn{4}{l}{\textbf{Different Demonstration Selection}}                                           \\ \hline
5-shot cluster-based planner  & 5-shot retrieval-based planner   & \textbf{51.6}        & 46.0                   \\
20-shot cluster-based planner & 20-shot retrieval-based planner  & \textbf{67.2}        & 32.0                   \\ \hline
\multicolumn{4}{l}{\textbf{Different Narrative Generation Method}}                                       \\ \hline
0-shot planner        & 0-shot without planner        & \textbf{76.7}        & 20.9                 \\
5-shot cluster-based planner &  5-shot cluster-based without planner & \textbf{88.2}        & 11.0                   \\
5-shot retrieval-based planner & 5-shot retrieval-based without planner & \textbf{70.6}        & 29.4                 \\ \hline
\end{tabular}
\caption{\textbf{Ablation Study Result. Different Demonstration Number}: In the learning stage of EIPE-text, in-context learning based planner use different numbers of demonstrations. \textbf{Different Demonstration selection}: In-context learning based planner can implement different methods, such as clustering or retrieving items related to the input topic, to select demonstrations. \textbf{Different Narrative Generation Method}: In addition to being able to generate narratives using EIPE-text. Narrative can also be generated in one step by simply combining several narratives as demonstrations without planner giving an input topic.}
\label{tab:5}
\end{table*}

\begin{table*}[]
\small
\centering
\renewcommand{\arraystretch}{1.3}
\begin{tabular}{c|ccc|cc|cc}
\hline
\multirow{2}{*}{metric} &
  \multicolumn{3}{c|}{operation} &
  \multicolumn{2}{c|}{difference before and after} &
  \multicolumn{2}{c}{epochs and question numbers} \\ \cline{2-8} 
 &
  \textbf{add} &
  \textbf{modify} &
  \textbf{adjust} &
  \textbf{all nodes} &
  \textbf{secondary nodes} &
  \textbf{average epoch} &
  \textbf{average questions} \\ \hline
num &
  8.26 &
  3.22 &
  2.25 &
  11.41 &
  0.25 &
  2.98 &
  35.71 \\ \hline
\end{tabular}

\caption{Iterative Refinement Metric}
\label{tab:6}
\end{table*}

\section{Analysis}
In this section, we explore the key aspects of designing an effective planner and provide an experimental analysis of the effectiveness of the plan refinement process.

\subsection{Ablation study of in-context learning based planner}
Our investigation centers around two fundamental questions: (1) How does the demonstration selection algorithm impact the performance of our planner? (2) What effect does the number of demonstration examples have on the planner's performance?

To address these questions, we designed experiments where we compared various planner configurations, including (1) \textbf{n-shot cluster-based planner}: this configuration utilizes a cluster-based approach to select n demonstration examples. (2) \textbf{n-shot retrieval-based planner}: in contrast, this configuration employs a retrieval-based method to select n demonstration examples.

\begin{figure}[htbp]
    \centering
    \includegraphics[width=8cm]{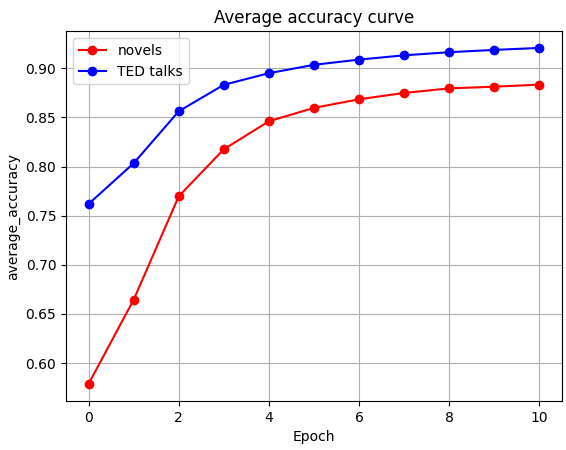}
    \caption{Average accuracy curve of iterative refinement process.}
    \label{fig:5}
\end{figure}

\textbf{Using clustering to select more demonstrations leads to better results.} We show the results in table \ref{tab:5}. In the comparison between the 20-shot cluster-based planner and the 5-shot cluster-based planner, the 20-shot cluster-based planner outperforms the 5-shot cluster-based planner with a win ratio of 70.9\% versus 26.8\%. This suggests that using more demonstration examples leads to better planner performance. In addition, as the plan length we use is shorter than full narrative, we can use more plans as demonstrations within context window. When comparing the 5-shot cluster-based planner and the 5-shot retrieval-based planner, the clustering-based method for selecting demonstration examples appears to be slightly more effective. This trend is more pronounced when looking at the comparison between the 20-shot cluster-based planner and the 20-shot retrieval-based planner. The 20-shot cluster-based planner significantly outperforms the retrieval-based planner, with a win ratio of 67.2\% versus 32.0\%. This suggests that using clustering for selection is considerably more effective than relying on retrieval-based methods.

\subsection{Comparison between hierarchical generation and non-hierarchical}
To investigate the impact of narrative generation methods on the performance of our planner, we compared hierarchical generation with non-hierarchical methods. 

We experiment with non-hierarchical generation including configurations:
(1) \textbf{0-shot without planner}: generate full narrative directly in one step. (2) \textbf{n-shot cluster-based without planner}: select n demonstrations using a cluster-based approach and generate a full narrative using these demonstrations. (3) \textbf{n-shot retrieval-based without planner}: similar to previous setting, instead, we rely on a retrieval-based approach to select demonstrations.

\textbf{Hierarchical generation is effective compared with non-hierarchical}. We show the results in table \ref{tab:5}. The 0-shot planner, significantly outperforms 0-shot without planner, achieving a win ratio of 76.7\% versus 20.9\%. Moreover, similar trends can be found in 5-shot setting with 88.2\% versus 11.0\% and 70.6\% versus 29.4\%.

\subsection{Effectiveness of the plan refinement process} 
In addition, we also want to know whether self-refinement can be effectively refined and the reasons behind its convergence.
\paragraph{Fast Convergence with Self-Refinement} We can see from the table \ref{tab:6} that our framework can converge in an average of 2.98 epochs, which is actually very fast and it is hard to converge without using self-refinement. The average accuracy curve of iterative refinement process is shown in figure \ref{fig:5}. 
\paragraph{Iterative Plan Refinement Ensures Alignment} The refined plan contains three operations, we monitor the number of three operations in the process. In addition, since we organize the plan into a tree structure, we also record the change in the number of nodes in the tree and the change in the number of secondary nodes (children of the root node) throughout the process. As can be seen from table \ref{tab:6}, the average add, modify and adjust operations occur 8.26 times, 3.22 times, and 2.25 times respectively. The average number of nodes increase by 11.41. We can clearly see these changes in figure \ref{fig:plan_comparison} (for more detail in appendix \ref{appendix:plan_extract}). This indicates that in plan refinement process, it does not simply add nodes. Instead, it can accurately modify relevant parts and adjust structure according to the question answering. Thus, these three operations ensure the alignment between the plan and the original narrative.

\subsection{Case study of in-context learning based plan generation}
Relying solely on comprehensive narratives for learning can often lead to missing finer details. Narratives are typically dense with information, posing challenges for models to pinpoint and retain critical elements. Furthermore, methods that learn from complete narratives are usually computationally expensive and time demanding. On the other hand, when using in-context learning with plans, models can more adeptly identify and relate to relevant information within each contextual segment. This technique not only ensures that key details aren't overlooked but also streamlines the learning process regarding the text's semantic framework, ultimately conserving computational resources. We show an example of 1-shot in Appendix \ref{appendix:plan_generate}, from which we can see that the generated plan is not only coherent but also retains the salient features of the demonstration, while effectively addressing the topic query.

\section{Related Work}
\paragraph{Long-form Narrative Text Generation}
As for long-form narrative text generation, recent studies tackle this from the following perspectives: appending the generated prefix to the encoder \cite{shao-etal-2017-generating}, while newer models like \citep{guan-etal-2021-long} focus on capturing sentence and discourse-level coherence, and DiscoDVT by \citep{ji-huang-2021-discodvt} leverages discrete variational Transformers to enhance long-range coherence in generated texts. Another type of work adopts the plan-and-write strategy \citep{fan-etal-2018-hierarchical}. In particular, there has been extensive exploration of story planning \citep{yao2019plan,fan-etal-2019-strategies,goldfarb-tarrant-etal-2020-content}. A hierarchical story generation system with recursive prompting and revision was proposed by \citet{yang-etal-2022-re3}. And the current state-of-the-art work recurrentGPT \citep{zhou2023recurrentGPT}, which uses large language model (LLM) such as ChatGPT and uses natural language to simulate the Long Short-Term Memory mechanism in an LSTM. The current plan results from these methods are not satisfactory. Instead, we use LLM to automatically mine the plan and train a good planner to achieve good results. Furthermore, from the plan to the full text, our methods and theirs are complementary and can be combined to achieve better results.
\paragraph{Human-AI Co-writing}
Human-AI co-writing systems have been developing at the intersection of NLP and human-computer interaction (HCI) fields, such as Wordcraft \citep{yuan2022wordcraft}, TaleBrush \citep{chung2022talebrush}, CoAuthor \citep{lee2022coauthor} and Dramatron \citep{mirowski2023co}. These works explore the possibilities of using LLM as a writing assistant to humans. Our work generates an explicit plan, which can be easily provided for human review and modification, making human-AI co-writing easier.

\section{Conclusions}
EIPE-text represents a significant step forward in the field of long-form narrative text generation, addressing the challenges of coherence and structure over extended pieces of text. With its ability to generate high-quality long-form narratives and aid human writers, EIPE-text opens up new possibilities for leveraging the capabilities of LLMs in creative and expressive writing tasks. Future research could explore further applications and extensions of EIPE-text in various domains, advancing the state of the art in automated text generation.
\section{Limitations}
During plan extraction stage, the two steps of QA-pairs generation and questions answering largely depend on LLM's own reasoning capability, so this method can only produce ideal results on models with strong reasoning capability (GPT4, Claude, etc.). Otherwise, it may lead to the refinement process failing to converge. Our framework is a data-driven approach, so it does not improve the OOD performance.


\bibliography{anthology,custom}

\appendix

\section{Prompts}
\label{sec:appendixa}
\subsection{Prompt of Sketching Plan}
\label{sec:plan_sketching_prompt}
Distill the salient information and thematic flow from the original article into a tree-like text representation of a mind map in the following format:

\begin{verbatim}
TOPIC
  - Main Topic
      - Sub Topic
          - Sub-Sub Topic
          - Sub-Sub Topic
  ...
  - Main Topic
      - Sub Topic
      - Sub Topic
\end{verbatim}

\subsection{Prompt of QA-pairs Generation Guideline}
\label{sec:qa_generation_guideline}
Based on the content of the article, generate several multiple-choice questions and corresponding answers:

\begin{enumerate}
  \item Not too detailed
  \item Focus on the logic of the article
  \item Deep understanding of the article after answering these questions
  \item Each question must have 4 options: A, B, C, D.
  \item For each question, there might be more than one correct answer, identify all correct answers separated by ";"
  \item Questions should reflect the structure of the article.
  \item Questions should include three types: what, why, how.
  \item Provide related main ideas in the article for each question.
  \item Avoid options like "All of the above" or "None of the above"; use "A;B;C" format.
\end{enumerate}
These questions are generated based on the article's content and the author's opinion, not my opinion.

\section{Experiment Details}
\subsection{Modification for recurrentGPT}
\label{sec:appendixb}
\begin{figure}[]
    \centering
    \includegraphics[width=8cm]{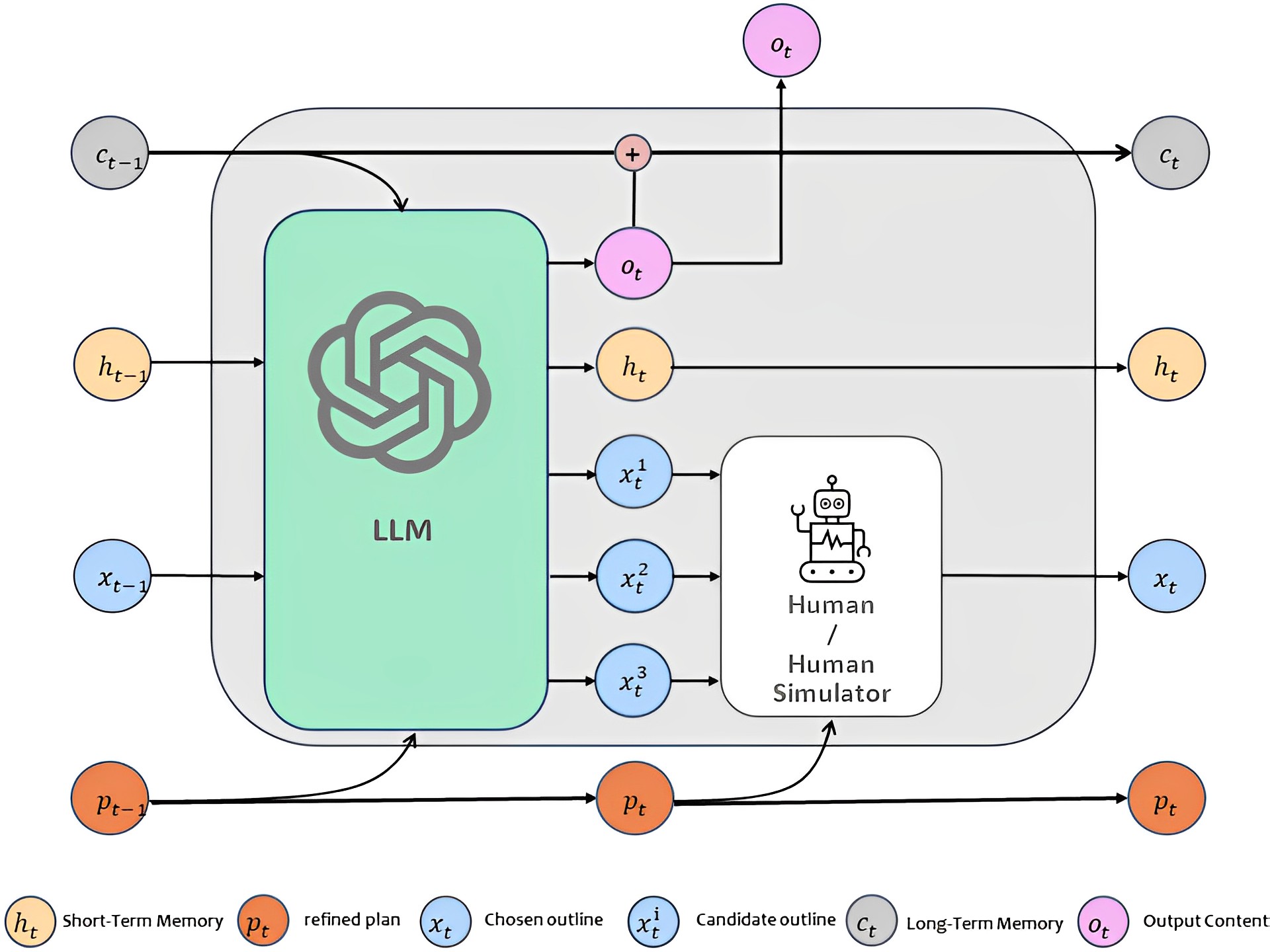}
    \caption{Illustration of the improvement of recurrentGPT. recurrentGPT uses natural language simulating an LSTM, we insert additional memory to main the long-term memory in LSTM}
    \label{fig:2}
\end{figure}

The way to improve recurrentGPT. recurrentGPT is prone to loss of global memory just as RNN. And we also find that the long-term memory in recurrentGPT is not exactly long-term memory. To compensate for this, we can insert the generated plan as additional memory to recurrentGPT, as shown in figure \ref{fig:2}. At each time step ('t'), recurrentGPT operates on a dual input system: the paragraph produced in the preceding step and a concise yet directive instruction for the subsequent paragraph. A crucial aspect is the integration of the model's long-term memory, which acts as a repository for storing previously generated summaries, and importantly, it can retrieve these summaries through semantic search, with the ability to store them on external hard drives. Simultaneously, the system actively maintains a short-term memory, responsible for encapsulating key information from recent time steps, a repository that gets updated consistently as the process unfolds. Crucially, the "generated plan," a newly introduced memory, becomes an integral part of this intricate orchestration. When the components converge, they create a coherent prompt that triggers the backbone language model, aptly dubbed the "backbone LLM," to undertake its primary task: generating a fresh paragraph while simultaneously outlining a succinct plan for the forthcoming paragraph. What's truly remarkable is how this "generated plan" seamlessly merges with the process, as it is not only updated in each time step but also contributes to enriching the long-term memory. This meticulous integration ensures continuity and coherence throughout the sequence, forging a recurrent mechanism that drives the generation process forward, where the "generated plan" plays a pivotal role in shaping the narrative's development.

\subsection{Use K-Means Get Demonstrations}
Using text-embeddings-ada-002 directly to convert a text format plan to embedding for clustering is not varied. We have adopted a combination approach of K-means and LLM. To be specific, we use LLM, according to prompt "Without loss of generality, list distinctive characteristics of this exemplar that establish it as an effective paradigm for designing {genre}. no explanation is needed." to get characteristics of plans. Then these characteristics are converted into embedding before clustering. After we set the number of clusters k, we can get k clustering centers, and we use these centroid as final demonstrations.

\section{Automatic Evaluation}
\label{sec:appendixc}
\subsection{Novel Automatic Evaluation}
\label{sec:appendixc1}
In this task, you will be presented with two novels side-by-side and asked to evaluate them based on three metrics: Coherence, Interestingness, Relevance. Your task is to determine which novel is better for each metric or indicate if both novels are indistinguishable.

- Coherent: A coherent novel follows a logical and consistent plot-line without significant gaps or inconsistencies.

- Interesting: An interesting novel captivates the reader's attention, engages them emotionally, and holds their interest throughout.

- Relevant. Faithful to the initial premise. The novel effectively aligns its plot, message, and writing with its initial premise, ensuring consistency and faithfulness to the core theme.

Based on these three aspects, make a decision on which novel is achieving the desired impact, in the manner like this:

[Scratch Pad]

Name:

`distinctive characteristics` and `elaborate` on them

Name:

`distinctive characteristics` and `elaborate` on them

[Reflection]

After evaluating both novels based on the criteria of `coherence`, `interestingness`, `relevance`, I have come to the following `thorough` conclusions:

  - Coherence:
  
  - Interestingness:
  
  - Relevance:

[Final Choice]: 

Coherence: Name;

Interestingness: Name;

Relevance: Name;

\subsection{TED Talks Automatic Evaluation}
\label{sec:appendixc2}
The coach's preference for evaluating the TED Talks can be summarized in the following spec:

  - Coherence: The coach will assess how well the TED Talk is structured and organized. This includes a clear introduction, logical flow of ideas, smooth transitions between points, and a strong conclusion. The talk should be easy to follow and understand, with a consistent theme throughout.
  
  - Interestingness: The coach will evaluate how engaging and captivating the TED Talk is for the audience. This includes the use of storytelling, anecdotes, and examples to illustrate points, as well as the speaker's ability to maintain the audience's attention and curiosity throughout the talk.
  
  - Relevance: The coach will consider the importance and significance of the topic being discussed in the TED Talk. The subject matter should be timely, relevant to current events or societal issues, and have a broad appeal to a diverse audience. The talk should also provide new insights or perspectives on the topic, rather than simply rehashing existing information.
  
  - Inspiration: The coach will assess the TED Talk's ability to inspire, motivate, and provoke thought in the audience. This includes the speaker's ability to convey passion and enthusiasm for the topic, as well as the presentation of innovative ideas, solutions, or calls to action that encourage the audience to think differently or take action in their own lives.

Based on these four aspects, the coach will make a decision on which TED Talk is stronger and more effective in achieving the desired impact on the audience, in the manner like this:

[Scratch Pad]

Name:

`distinctive characteristics` and `elaborate` on them

Name:

`distinctive characteristics` and `elaborate` on them

[Reflection]

After evaluating both TED Talks based on the criteria of `coherence`, `interestingness`, `relevance`, and `inspiration`, I have come to the following `thorough` conclusions:

  - Coherence:
  
  - Interestingness:
  
  - Relevance:
  
  - Inspiration:

[Final Choice]: Name

\section{Examples}
\subsection{Plan Extraction Example}
\begin{figure*}[h]
    \centering
    \includegraphics[page=5, width=\textwidth]{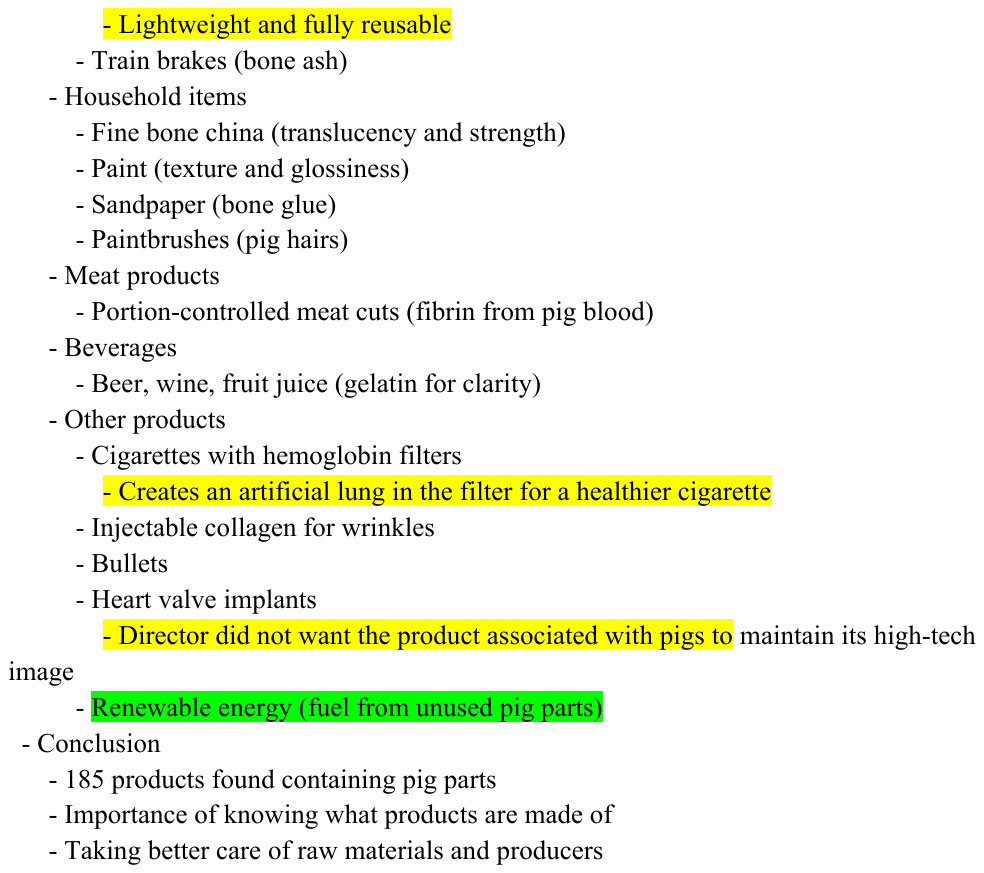}
\caption{The detailed process of an iteration in the plan extraction stage}
\label{fig:detail}
\end{figure*}

\begin{figure*}[!t]
    \centering
    \includegraphics[width=\linewidth]{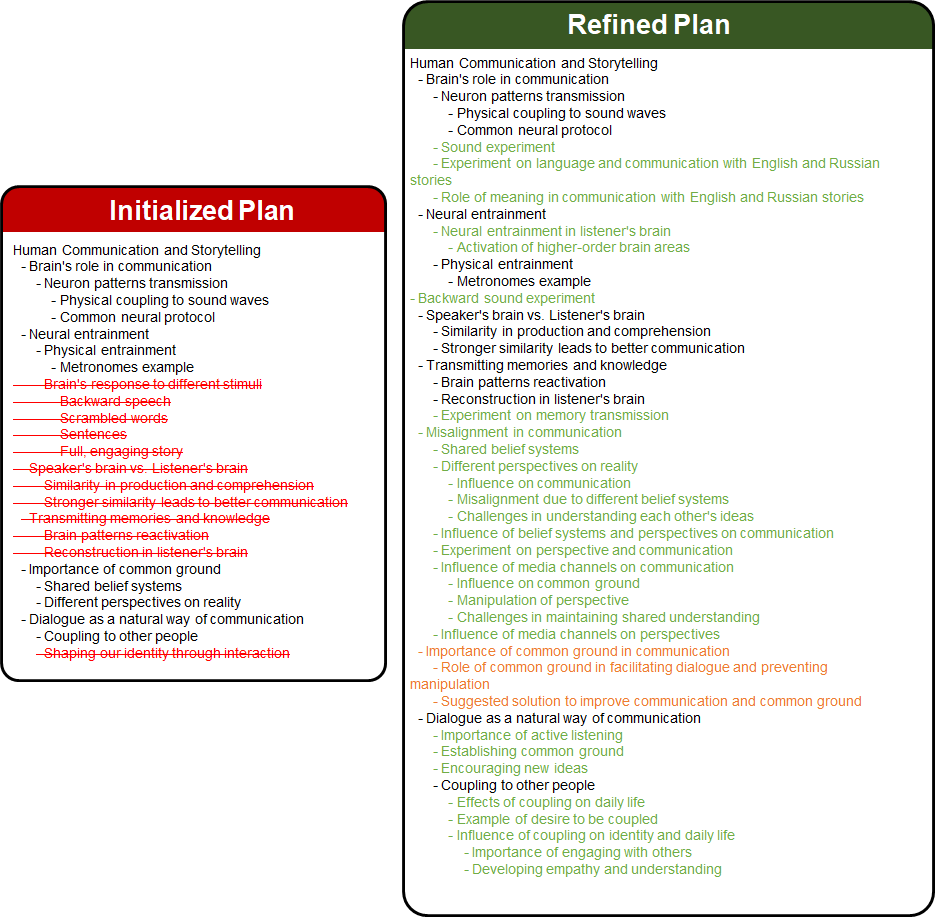}
    \caption{\textbf{Topic: This is your brain on communication}. Neuroscientist Uri Hasson researches the basis of human communication, and experiments from his lab reveal that even across different languages, our brains show similar activity, or become \"aligned,\" when we hear the same idea or story. This amazing neural mechanism allows us to transmit brain patterns, sharing memories and knowledge. \"We can communicate because we have a common code that presents meaning,\" Hasson says.}
    \label{fig:res1}
\end{figure*}

\begin{figure*}[!t]
    \centering
    \includegraphics[width=\linewidth]{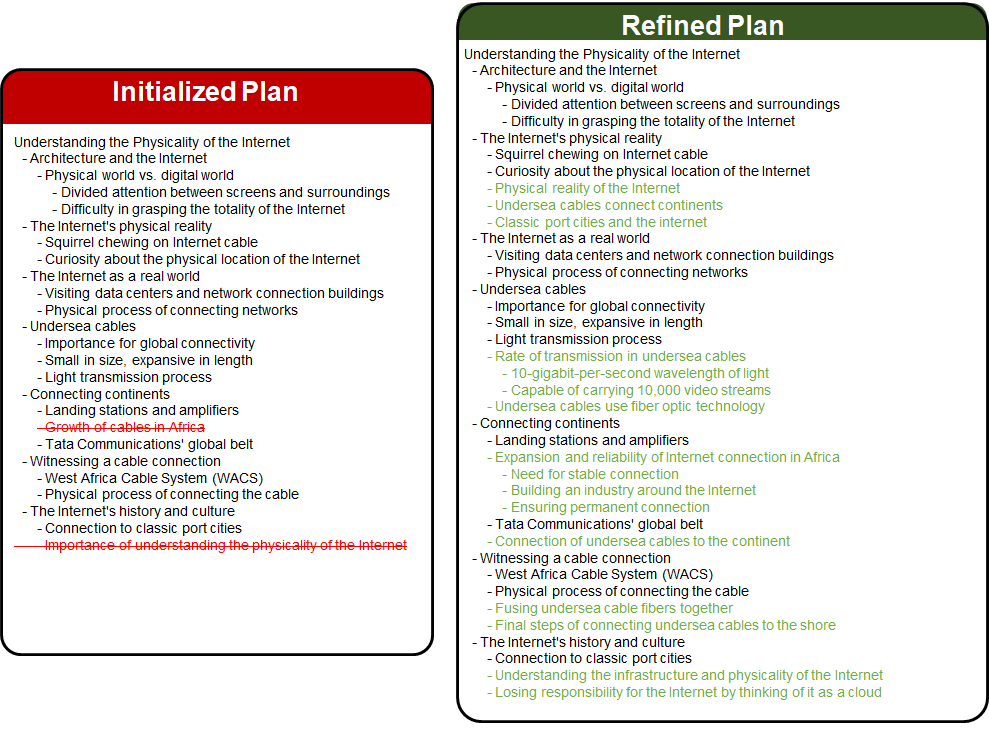}
    \caption{\textbf{Topic: Discover the physical side of the internet}. When a squirrel chewed through a cable and knocked him offline,  journalist Andrew Blum started wondering what the Internet was really made of. So he set out to go see it -- the underwater cables, secret switches and other physical bits that make up the net.}
    \label{fig:res2}
\end{figure*}

\begin{figure*}[!t]
    \centering
    \includegraphics[width=\linewidth]{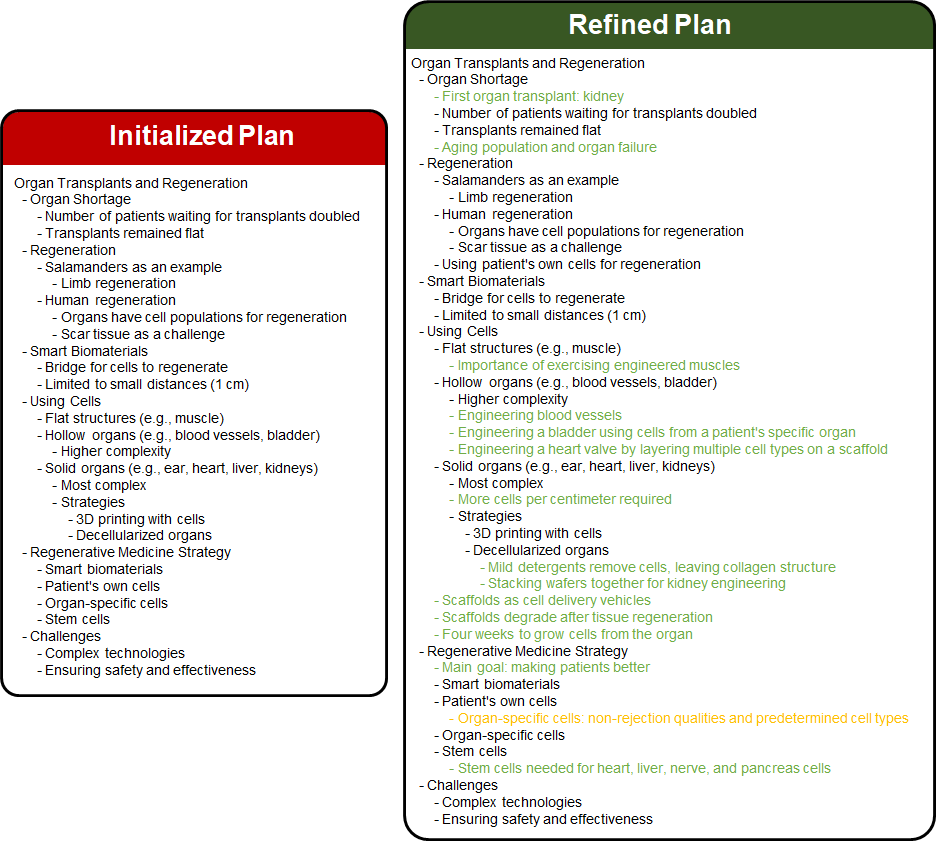}
    \caption{\textbf{Topic: Growing new organs}. Anthony Atala's state-of-the-art lab grows human organs -- from muscles to blood vessels to bladders, and more. At TEDMED, he shows footage of his bio-engineers working with some of its sci-fi gizmos, including an oven-like bioreactor (preheat to 98.6 F) and a machine that \"prints\" human tissue.}
    \label{fig:res3}
\end{figure*}

In this section, we show the detailed process of an iteration in the plan extraction stage in the figure \ref{fig:detail}. And the comparison of initialized plans and refined plans are also shown in figure \ref{fig:res1}, \ref{fig:res2}, \ref{fig:res3}
\label{appendix:plan_extract}
\begin{figure*}[h]
    \centering
    \includegraphics[page=1, width=\textwidth]{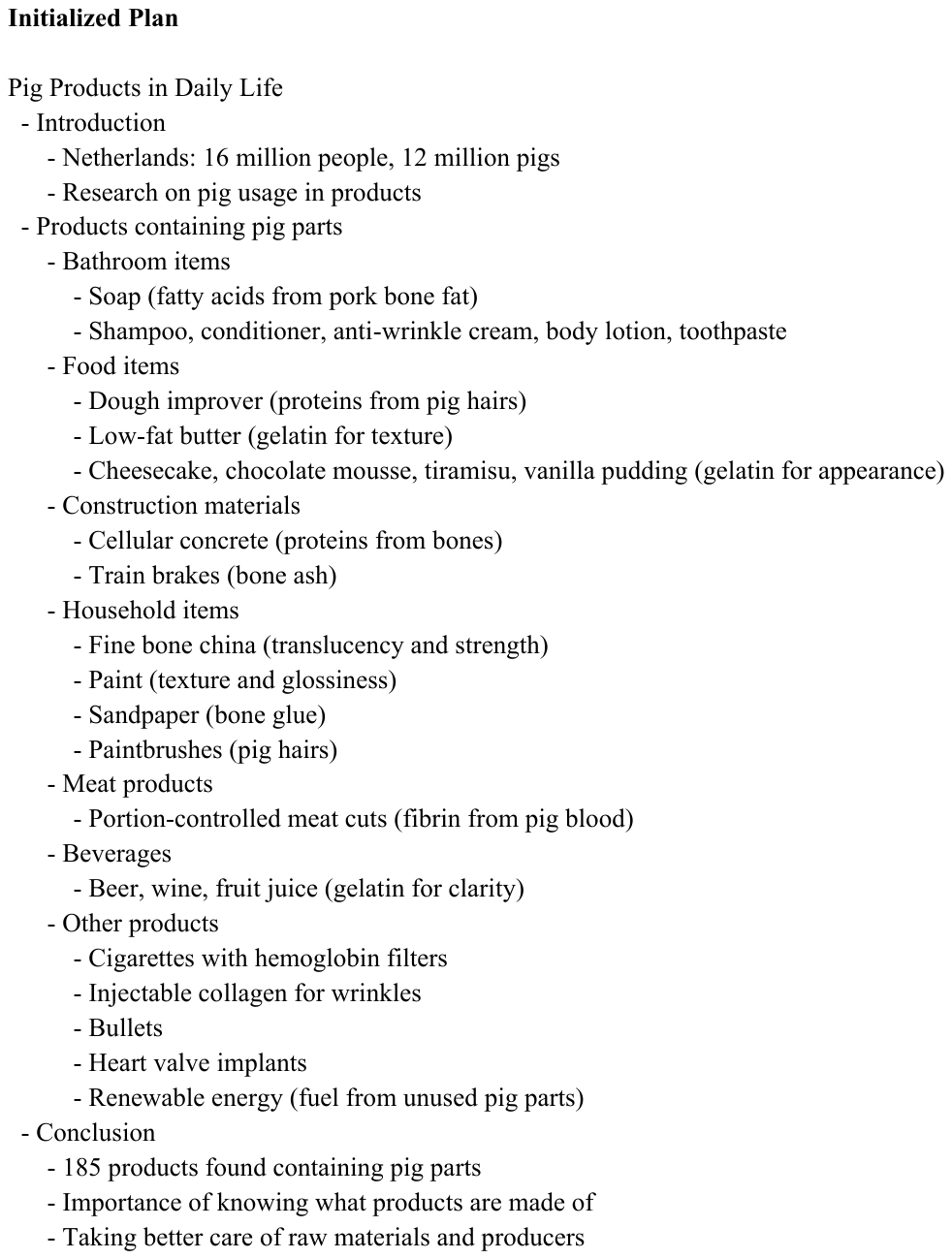}

\end{figure*}

\begin{figure*}[h]
    \centering
    \includegraphics[page=2, width=\linewidth]{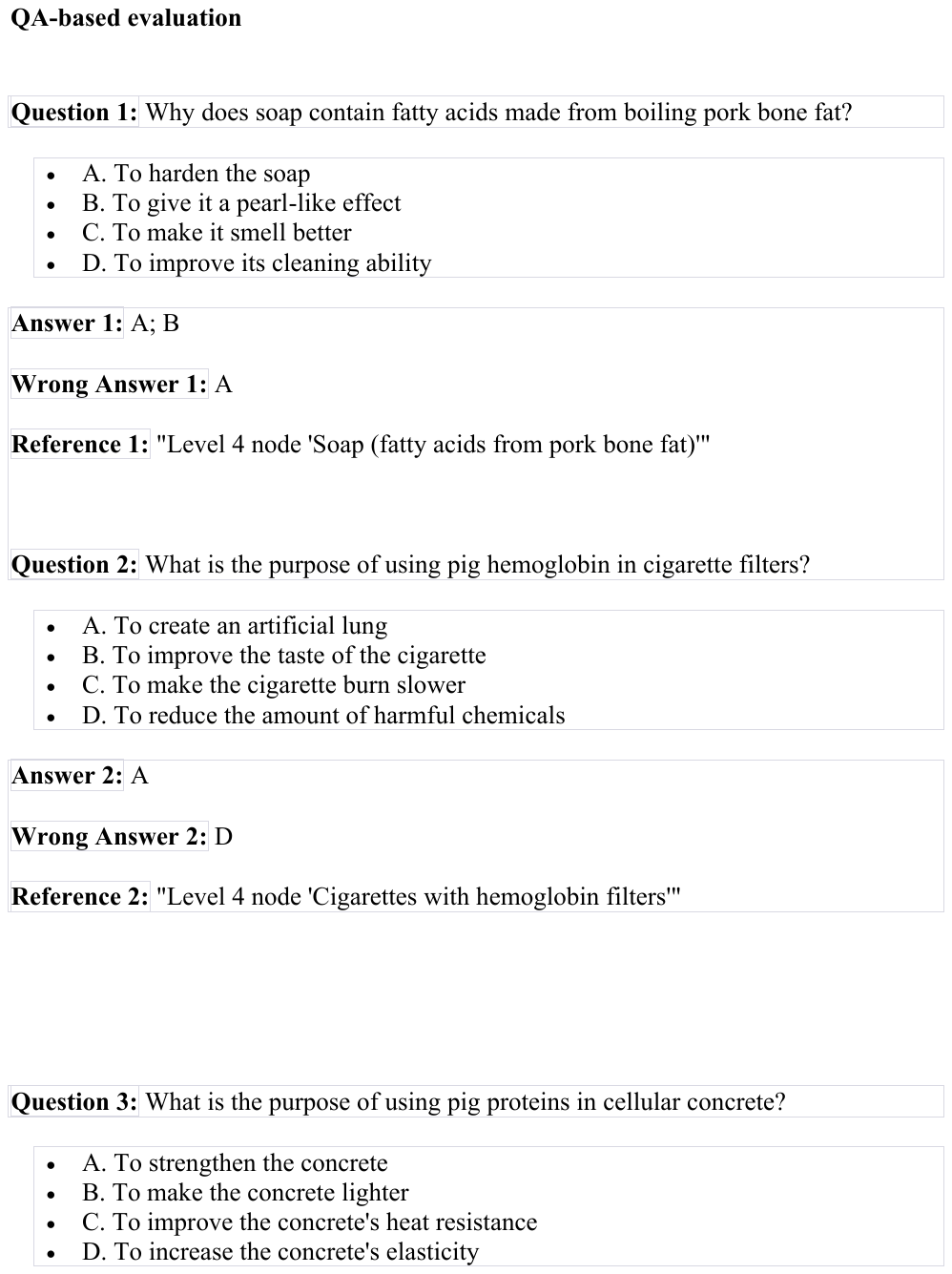}

\end{figure*}

\begin{figure*}[h]
    \centering
    \includegraphics[page=3, width=\textwidth]{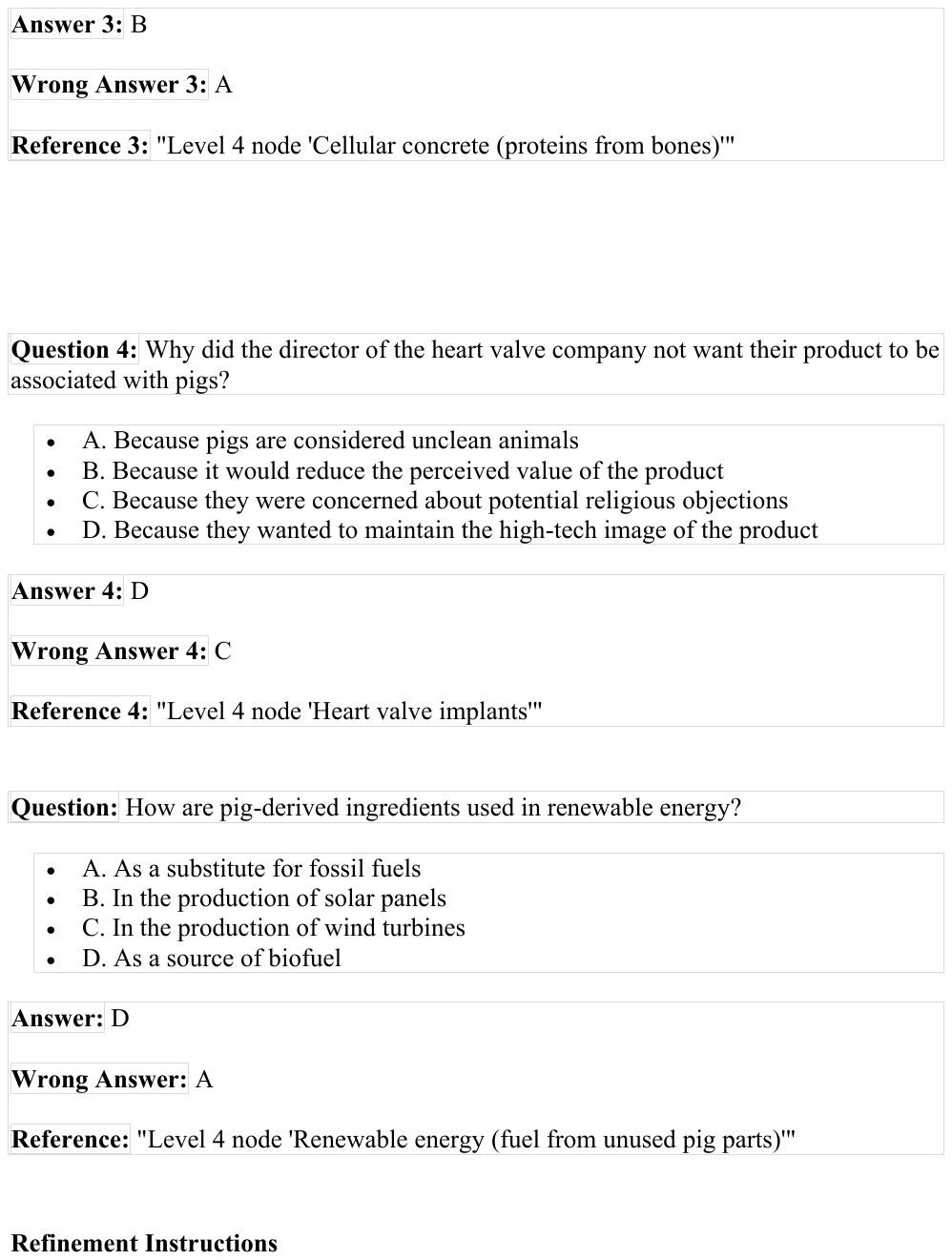}

\end{figure*}
\begin{figure*}[h]
    \centering
    \includegraphics[page=4, width=\textwidth]{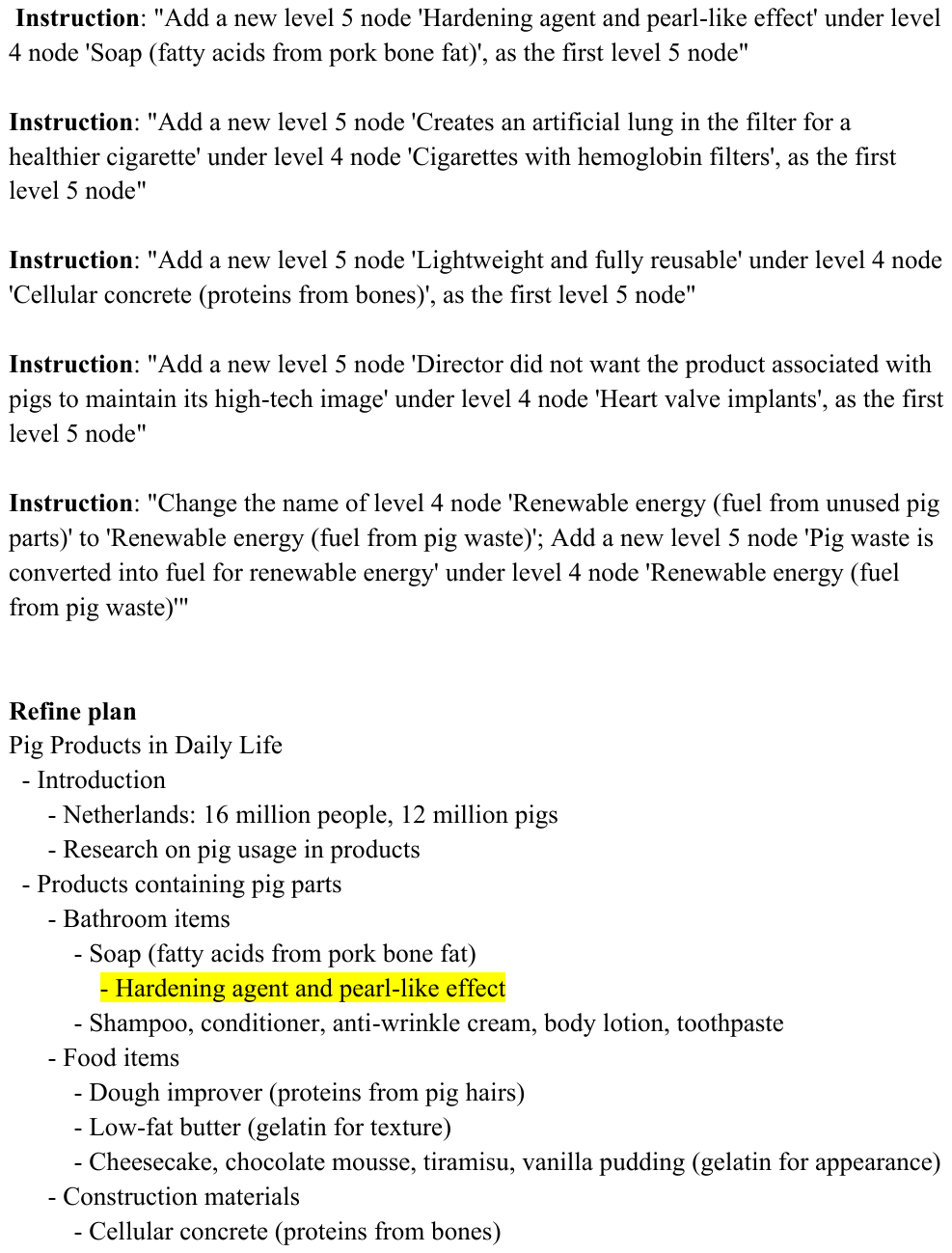}

\end{figure*}

\subsection{Plan Generation Example}
\label{appendix:plan_generate}
\begin{figure*}[tbp]
    \centering
    \includegraphics[width=\linewidth]{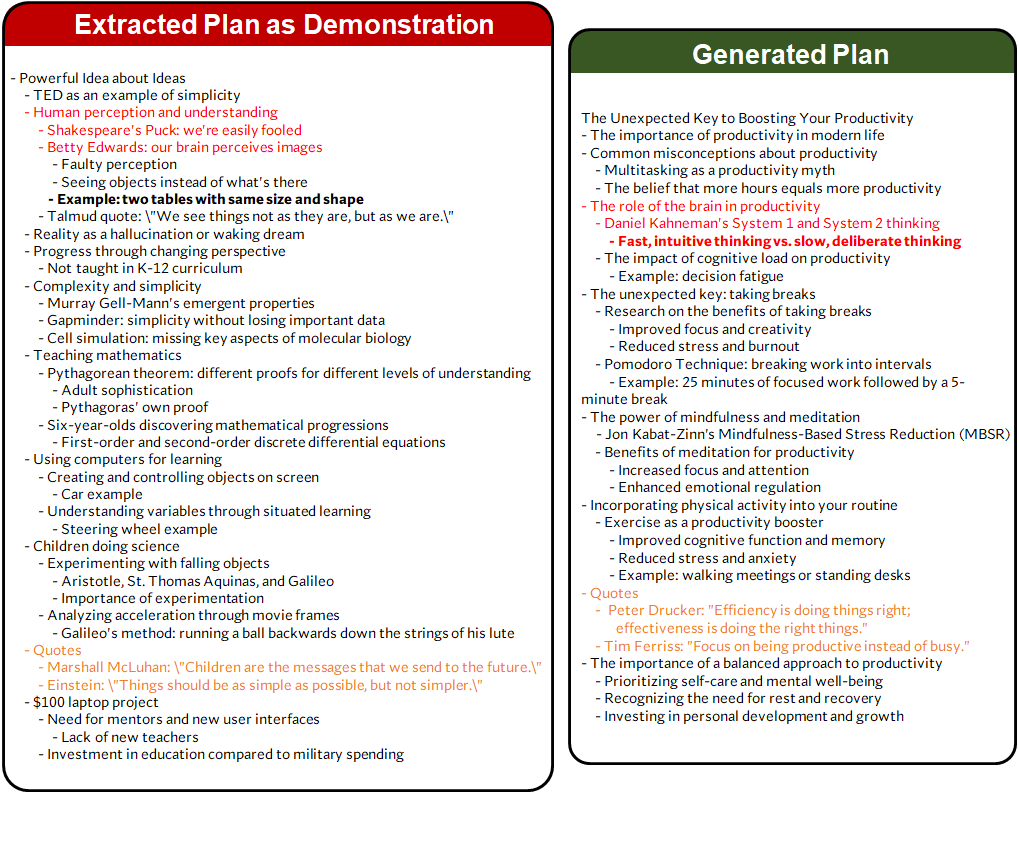}
    \caption{Example of one shot. }
    \label{fig:example_source}
\end{figure*}

In figure \ref{fig:example_source}, we observe several salient aspects of the demonstration on the left side: 1. the utilization of relatable examples and analogies. 2. the connection of various disciplines and concepts. 3. the incorporation of quotes from notable figures. These attributes are integrated into the generated plan. Notably, the generated plan contains relatable examples to substantiate its viewpoints and incorporates relevant quotes. When we transition to the generated plan on the right, it's evident that the plan incorporates these attributes seamlessly. For instance: 1. the mention of "Daniel Kahneman’s System 1 and System 2 thinking" in the plan mirrors the demonstration's theme on "Human perception and understanding.". 2. the outcome's emphasis on quotes is reflective of the demonstration's approach, incorporating wisdom from Peter Drucker and Tim Ferriss. We can observe that the generated plan not only maintains coherence but also preserves the key features of the demonstration, while effectively responding to the topic query.




\end{document}